\documentclass[10pt, a4paper]{article}
\usepackage{lrec2022} 

\usepackage{multibib} 
\newcites{languageresource}{Language Resources} 
\usepackage{graphicx}
\usepackage{tabularx}
\usepackage{booktabs}
\usepackage{bm}
\usepackage{soul} 
\usepackage{float}
\usepackage{tipa}
\usepackage[neverdecrease]{paralist}
\usepackage{comment}
\usepackage{multirow}

\usepackage{titlesec}
\titleformat{\section}{\normalfont\large\bfseries\center}{\thesection.}{1em}{}
\titleformat{\subsection}{\normalfont\SmallTitleFont\bfseries\raggedright}{\thesubsection.}{1em}{}
\titleformat{\subsubsection}{\normalfont\normalsize\bfseries\raggedright}{\thesubsubsection.}{1em}{}
\renewcommand\thesection{\arabic{section}}
\renewcommand\thesubsection{\thesection.\arabic{subsection}}
\renewcommand\thesubsubsection{\thesubsection.\arabic{subsubsection}}

\usepackage{epstopdf} 
\usepackage[utf8]{inputenc}

\usepackage{hyperref}
\usepackage{xstring}

\usepackage{color}

\newcommand{\word}[1]{\textit{#1}}
\newcommand{\defn}[1]{\textbf{#1}}

\usepackage{amsmath}
\usepackage{amsfonts}

\DeclareMathOperator{\similar}{sim}
\DeclareMathOperator{\lemma}{lemma}
\DeclareMathOperator{\emb}{emb}
\DeclareMathOperator{\metric}{proximity}
\DeclareMathOperator*{\argmax}{\arg\!\max}
\DeclareMathOperator{\bow}{bow}

\usepackage{cleveref}

\crefname{section}{\S}{\S\S}
\Crefname{section}{\S}{\S\S}
\crefname{table}{Table}{}
\crefname{figure}{Figure}{}
\crefname{algorithm}{Algorithm}{}
\crefname{equation}{eq.}{}
\crefname{appendix}{App.}{}
\crefname{prop}{Proposition}{}
\crefformat{section}{\S#2#1#3}

\usepackage{todonotes}

\title{Homonymy Information for English WordNet}

\name{Rowan Hall Maudslay~\;~~\;~Simone Teufel} 

\address{Dept.\ of Computer Science and Technology \\
University of Cambridge \\
         {\tt \{rh635,sht25\}@cam.ac.uk}\\}

\usepackage{enumitem}
\everypar{\looseness=-1}

\abstract{
A widely acknowledged shortcoming of WordNet is that it lacks a distinction between word meanings which are systematically related (polysemy), and those which are coincidental (homonymy). Several previous works have attempted to fill this gap, by inferring this information using computational methods.  We revisit this task, and exploit recent advances in language modelling to synthesise homonymy annotation for Princeton WordNet. Previous approaches treat the problem using clustering methods; by contrast, our method works by linking WordNet to the Oxford English Dictionary, which contains the information we need. To perform this alignment, we pair definitions based on their proximity in an embedding space produced by a Transformer model. Despite the simplicity of this approach, our best model attains an F1 of $.97$ on an evaluation set that we annotate. The outcome of our work is a high-quality homonymy annotation layer for Princeton WordNet, which we release.
 \\ \newline  }

\begin{document}

\maketitleabstract

\section{Introduction}

Words have multiple meanings that are related to each other in different ways. Meanings which are systematically related are said to exhibit \defn{polysemy}. One example of polysemy is the use of the same wordform to refer to a product or its producer (\citealp{pustejovsky1995generative}):
\setdefaultleftmargin{.9cm}{.68cm}{}{}{}{}
\begin{enumerate}[label=({\arabic*})]
\item \label{ex:alternation}
\begin{enumerate}[label={\alph*}.]
\itemsep0em 
    \item John spilled coffee on the \word{newspaper}.
    \item The \word{newspaper} fired its editor.
\end{enumerate}
\end{enumerate}
Aside from such highly productive alternation patterns, 
polysemy also includes semi-productive metaphorical extensions \cite{lakoff1980metaphors}:
\begin{enumerate}[resume*, start=2, label=({\arabic*})]
\item \label{ex:metaphor}
\begin{enumerate}[label={\alph*}.]
\itemsep0em 
    \item They \word{adopted} a child.
    \item The theory was rapidly \word{adopted}.
\end{enumerate}
\end{enumerate}
Polysemy exemplifies humans' ability to flexibly extend categories to cover new members, which is of significant interest to researchers in cognitive science (\citealp{lakoff1987women}).
These extensions include figurative uses, like in example \ref{ex:metaphor}.
The polysemisation of words also plays a key role in lexical evolution and semantic drift (e.g.\ \citealp{koch2016meaning}).

On the other hand, meanings of the same word which exhibit no systematic relation are described as instances of \defn{homonymy}.\footnote{This is sometimes called `incidental polysemy', which is contrasted with `systematic polysemy' (e.g.\ \citealp{pustejovsky1995generative}).} 
These associations are non-productive, and result instead from language change. Usually, this occurs when new word senses are borrowed from other languages, and can involve vowelshifts and similar transformations. 
For example, consider the English word \word{bank}: 
\begin{enumerate}[resume*, start=3, label=({\arabic*})]
\item \label{ex:bank}
\begin{enumerate}[label={\alph*}.]
\itemsep0em 
    \item I need to get money out from the \word{bank}.
    \item Let's sit by the river on the \word{bank}.
\end{enumerate}
\end{enumerate}
The financial sense has its origin in the romance languages, and the river-edge sense comes from Old Norse. 
Another example of homonymy happens when acronyms become conventionalised, and are ultimately lower cased (e.g.\ Personal Identification Number):
\begin{enumerate}[resume*, start=4, label=({\arabic*})]
\item \label{ex:acronym}
\begin{enumerate}[label={\alph*}.]
\itemsep0em 
    \item Put a \word{pin} in the hem of the fabric.
    \item Never share your credit card's \word{pin}.
\end{enumerate}
\end{enumerate}

Although homonymous meanings are not semantically related, their presence in a particular language is not random, and instead may serve a communicative function \cite{piantadosi2012communicative}. 

WordNet \citelanguageresource{miller1995wordnet} is a popular computational lexicon. 
In WordNet, concepts are represented as an equivalence class of wordforms associated with that concept, called synsets.
WordNet makes no distinction between polysemy and homonymy.
If it did, WordNet would have the potential to be an ideal repository for research into these phenomena.

Several researchers have acknowledged this shortcoming of WordNet, and have attempted to produce computational models to synthesise homonymy annotation for it (e.g.\ \citealp{utt-pado-2011-ontology}; \citealp{veale-2004-polysemy}; \citealp{freihat2013regular}).
We revisit this task using contemporary methods.
By exploiting large language models, we synthesise a high-quality annotation layer for distinguishing between polysemy and homonymy in the English Princeton WordNet. 

\begin{figure*}[t]
    \centering
    \includegraphics{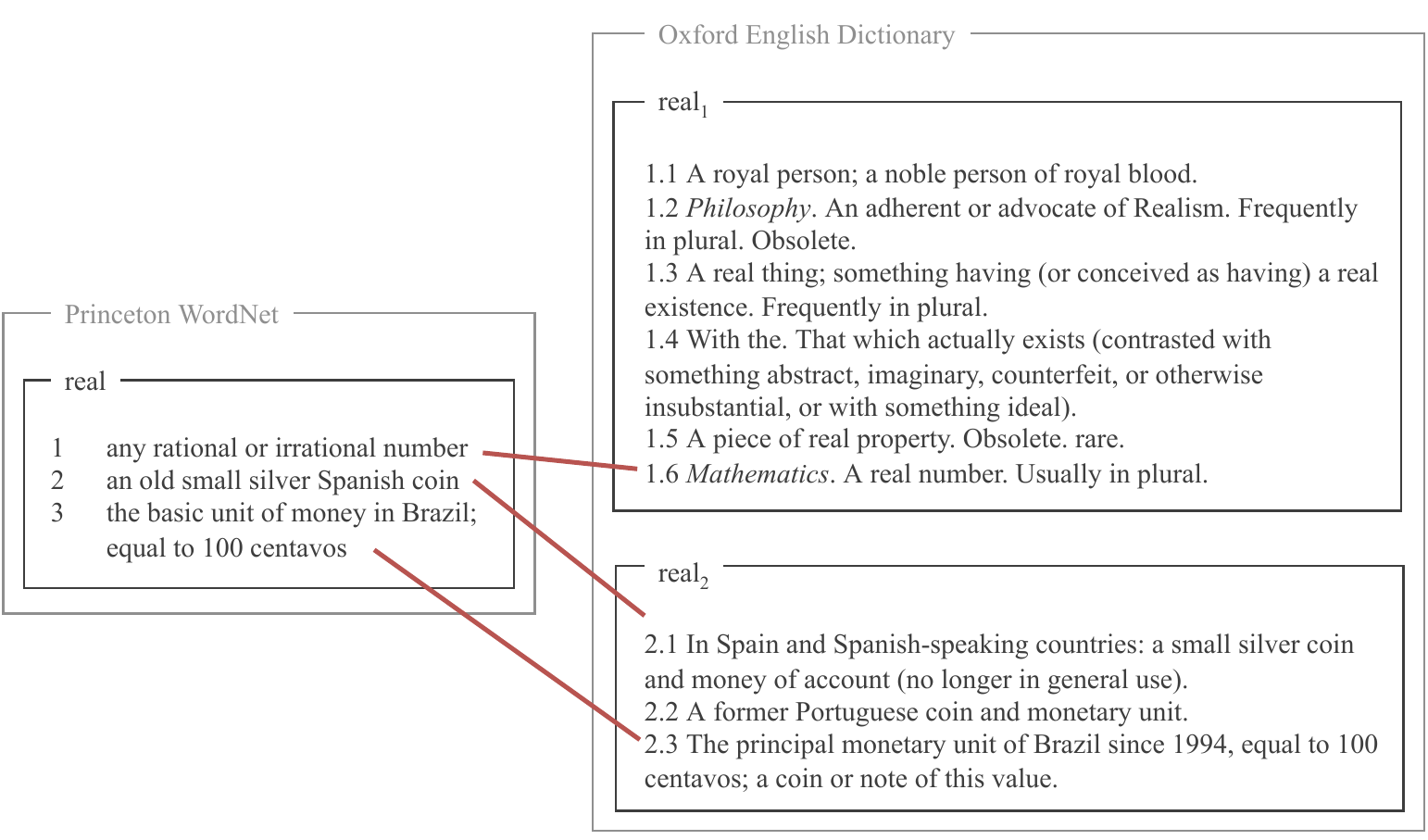}
    \caption{Noun definitions of the word \word{real} from the PWN (left) and the OED (right)}
    \label{fig:definitions}
\end{figure*}

More specifically, to identify homonyms in WordNet, we align it with the Oxford English Dictionary, a historical dictionary of English. 
In this dictionary, as a general principle in lexicography, a lemma is defined as a wordform plus all its polysemous senses. Homonymous wordforms are associated with multiple lemmas.
By aligning the senses in WordNet with corresponding senses in the Oxford English Dictionary, we can work out which lemma they belong to, and thus distinguish between senses which are related by polysemy (same lemma), and those related by homonymy (different lemmas). 
Previous works that attempted to identify homonymy in WordNet did so by clustering senses. 
An advantage of our linking approach is that figurative senses can be correctly identified as instances of polysemy, even though their meaning might differ radically from the literal sense they extend.

To align the dictionaries, we compute the sentence embeddings of each definition using various Transformer models \cite{vaswani2017attention}, and find the definition in the Oxford English Dictionary which is closest in embedding space to each WordNet definition. 
To evaluate the quality of the model, we annotate a small evaluation set of $196$ words ($554$ senses).
Despite the simplicity of our unsupervised method, it attains an F1-score of $.97$ on our evaluation set, indicating that our synthesised data is high quality.

\section{Background}

The \defn{Princeton WordNet} (PWN) is an English computational lexicon, which maps wordforms to concepts, which are called synsets \citelanguageresource{miller1995wordnet}. 
Synsets are associated with a definition and often some example sentences, and are also linked to each other in a semantic network (consisting primarily of \emph{is a} and \emph{has a} relations). 
Since its creation, several works have added additional annotation layers to the PWN (e.g.\ \citealp{mendes2001enriching}, \citealp{puscasu-mititelu-2008-annotation}, \citealp{amaro2006enriching}). 
In research on polysemy and homonymy, we often want to build rich representations of each sense, and the PWN is associated with useful resources for that. 
One set of resources links synsets with textual examples, e.g.\ SemCor \citelanguageresource{miller1994semcor} and the NTU-MC \citelanguageresource{tan-bond-2011-building}.
Other resources link synsets to images depicting the synset, e.g.\ ImageNet \citelanguageresource{deng2009imagenet} and BabelPic \citelanguageresource{calabrese-etal-2020-fatality}.

What the PWN lacks, however, is information which distinguishes homonymy from polysemy. 
Consider the word \word{real}, the noun senses of which are shown in \cref{fig:definitions}. In the PWN (left), the senses appear in a single group. 
In the \defn{Oxford English Dictionary} (OED), however, the senses are divided into two separate lemmas, real\textsubscript{1} and real\textsubscript{2} (right).\footnote{These are the lemmas that result following our homonymy identification procedure, which is detailed in \cref{sec:coarsening}.}
The OED is an authoritative English historical dictionary: unlike the PWN, which is a contemporary lexicon that shows a snapshot of current English usage, the OED maps each wordform to all known senses that it has ever had. 
Senses in the same lemma have the same etymology and pronunciation, and are likely derived from each other, i.e.\ they are polysemous. 
Senses in different lemmas likely bare no systematic relation, i.e.\ they are homonymous. The word \word{real} exhibits homonymy, but the PWN does not encode this information.

The problem of separating homonymy from polysemy in the PWN has been recognised, and several works have attempted to address it. 
Because manually annotating this information for all of the PWN would be expensive, previous approaches have synthesised the data using computational methods (e.g.\ \citealp{utt-pado-2011-ontology}, \citealp{veale-2004-polysemy}, \citealp{freihat2013regular}).
These previous works all adopt a similarity-driven clustering approach to separate homonymy from polysemy.
The problem with this approach is that some polysemous senses appear ``further apart'' in semantic space than homonyms. 
For example, two polysemous senses connected by metaphor are often extremely different on the surface (e.g.\ the \word{body} of a human v.\ the \word{body} of a guitar), and so are easily confused with homonymy even though they are related.

To ensure that instances figurative polysemy are not incorrectly labelled as homonymy, we use etymological information for the identification of homonyms. 
More specifically, we align WordNet with the OED (red lines in \cref{fig:definitions}).
Our work is most similar to \newcite{navigli-2006-meaningful}), who also aligned the PWN with the OED to cluster PWN senses.
However, while their clustering was produced for the purpose of Word Sense Disambiguation (WSD), we do so for the purpose of research into polysemy and homonymy; because of these research aims, we coarsen the OED lemmas, as outlined in \cref{sec:coarsening}.

The data synthesised by \newcite{navigli-2006-meaningful}), originally released for a 2007 shared task \cite{navigli-etal-2007-semeval}, clusters WordNet~2.1 senses. Since the early time of this work, many new methodologies for dictionary alignment have emerged.
Several works have aligned WordNet with other resources, for example Wiktionary and Wikipedia \citelanguageresource{miller-gurevych-2014-wordnet,meyer-gurevych-2011-psycholinguists,mccrae2012integrating,navigli2021ten}. 
Recently, a shared task was held on supervised monolingual dictionary alignment \cite{globalex-2020-2020}; in the English subtask, models were tasked with aligning the PWN with a publicly accessible version of the Webster’s dictionary from 1913 \citelanguageresource{ahmadi-etal-2020-multilingual}.
All models participating in the subtask use a Transformer model \cite{vaswani2017attention} in some form. 
Transformer models are sentence encoders, which produce embeddings for each input token. 
In our work, we revisit \newcite{navigli-2006-meaningful}), and use Transformer models to produce a high quality alignment for WordNet~3.1.

Finally, we note that while a resource called `Etymological Wordnet' already exists \citelanguageresource{de-melo-2014-etymological}, this resource is in fact unrelated to the WordNet project \citelanguageresource{miller1995wordnet}: it is an automatically extracted database of wordform derivations from Wiktionary. 

\section{Processing the OED}

In this section, we describe how we extract homonymy data from the OED (\cref{sec:coarsening}), and then how we collect data to evaluate model performance (\cref{sec:eval-data}).

\subsection{Extracting Homonyms from the OED} \label{sec:coarsening}

For every wordform with multiple senses in the PWN, we retrieve the corresponding lemmas from the OED.\footnote{Content provided by OED Researcher API, 2022.} 
Lemmas in the OED have etymology data associated with them, in the form of the language family of origin. 
Depending on the records available, some lemmas are annotated with more broad family information (e.g.\ Italic), while others have more fine grained information (e.g.\ French). 
Some have unknown origin. 
Because of this, sometimes it is ambiguous as to whether two lemmas are in fact related.

In these cases, we have to make a decision. 
We could either divide PWN senses into the lemmas as they are presented in the OED (and risk splitting polysemous senses into different lemmas), or we could merge lemmas together (and risk putting hymonymous senses into the same lemma). 
We choose to do that latter, because for research in these areas it is preferable to overestimate polysemy and underestimate hymonymy:
if two polysemous senses were wrongly separated into different lemmas, this would provide a wrong gold standard for any model of polysemisation. 

Our procedure for merging OED lemmas is as follows. 
Some lemmas are marked as being derived from others; in this case, we merge them with the lemma they are derived from. 
If there are multiple lemmas which have the same etymological derivation, we merge them. 
If one lemma's derivation is a subclass of another's (as with French v.\ Italic), we merge them. 
The exception to these merges is when a derivation is labelled as being the conventionalisation of an acronym; we leave these in their own lemma. 
Finally, if a lemma for a particular wordform has unknown etymology, we exclude that wordform (and thus assume that all its senses are polysemous).

\subsection{Annotating an Evaluation Set} \label{sec:eval-data}

\paragraph{Sampling Data} We sample wordform--part-of-speech combinations, which meet the following criteria:
\begin{itemize}
    \item have at least two senses in the PWN;
    \item have at least two lemmas in the OED (following our coarsening procedure, \cref{sec:coarsening}), and further, that at least two of these lemmas have at least two senses (to avoid severely imbalanced lemmas);
    \item have a maximum of $15$ senses overall in the OED (to reduce the cognitive load on annotators)
\end{itemize}
Following the above procedure, we sample $100$ wordform--part-of-speech combinations. These combinations had an average of $2.18$ lemma options in the OED, and yielded $286$ PWN senses.

\paragraph{Annotation Procedure} We need to collect a mapping of PWN senses to OED lemmas. However, as we will see in \cref{sec:method}, the models we study work by aligning PWN senses to OED senses. Although this is not our primary concern, it would be interesting to also evaluate how well models perform at this finer granularity of analysis. Because of this, we decide to ask annotators to assign each PWN sense to a single OED sense, from which we can trivially recover the sense-to-lemma mapping which is our main interest.
More specifically, we ask annotators to go through each word, and assign each PWN senses to a single OED sense. If there are multiple OED senses which would work, we ask them to select the best one. If there is no OED sense to align a PWN sense to, but there is an OED sense which is more broad and would include that PWN sense, we ask them to select that OED sense. If there is still not an appropriate OED sense, annotators have a choice. If they think the PWN sense is closely related to OED senses in a particular lemma, they assign the PWN sense to that lemma. Otherwise, if they think that the PWN sense is a different lemma, not contained in the OED, they leave it unassigned. 

\paragraph{Recovering Lemma Assignments} With the fine-grained sense-to-sense alignment which our annotators produce, we can reconstruct the sense-to-lemma mapping trivially. For each PWN sense that is aligned with an OED sense, we simply take the lemma that that OED sense is contained within in the OED. 

\paragraph{Statistics and Agreement} 
Two native British English speakers performed our annotation task. 
It is however not possible to report agreement in terms of chance-corrected Inter-Annotator Agreement (IAA) for a dictionary alignment task, because the number of possible categories that an item is assigned to varies depending on the wordform; we therefore report raw agreement. 
Both annotators gave the same lemma assignment $97.6\%$ of the time, and the same sense assignment $80.4\%$ of the time. 
$1.0\%$ of the time, at least one annotator judged that no lemma existed for a PWN sense.
$9.1\%$ of the time, at least one annotator judged that none of the fine-grained senses was appropriate, but that an appropriate lemma existed.
For comparability to similar tasks, we follow Ahmadi et al.\ (2020), and also compute IAA in terms of $\kappa$. 
Ahmadi et al.\ do this by treating each possible pair of senses (one from each dictionary) as a binary datapoint, which could be labelled $0$ if they were not aligned, or $1$ if they were . 
(However, we note that this method is problematic, as it overestimates agreement. This is because computations of $\kappa$ assume that each datapoint is independent, and under this formulation many of the datapoints are counted as agreement although they are simply a consequence of other decisions.)
Under these conditions, we find  $\kappa{=}0.96$ ($N{=}909$, $k{=}2$, $n{=}2$) for the lemma assignments, and $\kappa{=}0.79$ ($N{=}3{,}396$, $k{=}2$, $n{=}2$) for the sense assignments.
The high agreement is in line with previous work; \newcite{navigli-2006-meaningful}) found $\kappa=0.85$ for sense-level alignment between the PWN and the OED (although it is unclear how they performed this computation).

\paragraph{Evaluation Data} Having shown that our annotation procedure yielded high agreement, one annotator continued the annotation task for more examples, and labelled $96$ more wordforms which met the above criteria. This yields a final annotated set consisting of $196$ wordform--part-of-speech combinations covering $544$ PWN senses, which we will use to evaluate model performance, \cref{sec:eval}. In this final evaluation data, $1.3\%$ of PWN senses are not assigned to an OED lemma.

\begin{figure}[t]
    \centering
    \includegraphics{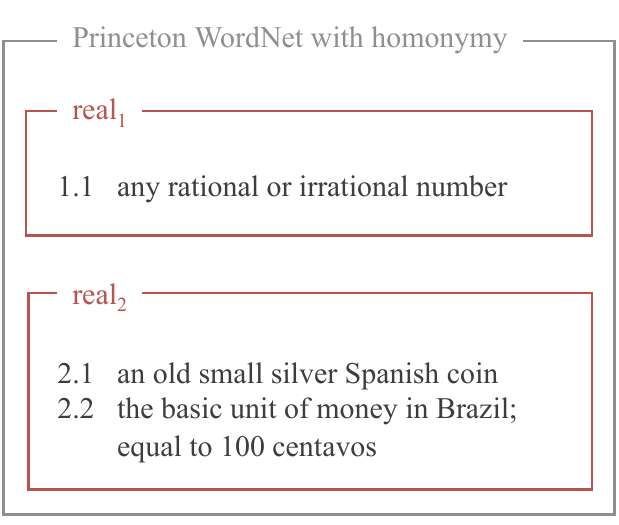}
    \caption{Our output annotation for the word \word{real}}
    \label{fig:definitions-split}
\end{figure}

\section{Method} \label{sec:method}

Our goal is to split homonymous PWN senses into separate lemmas (\cref{fig:definitions-split}). 
To achieve this, we align the PWN with the OED, in which senses are grouped according to their etymological derivation. 
Our method is a simple unsupervised approach, which pairs each definition from the PWN with the definition in the OED that it is closest to it in embedding space.

Let $\mathcal{S}$ be a set of all senses, which we take as string definitions. 
Let $\mathcal{S}_\textrm{OED}^{w}\subseteq\mathcal{S}$ denote the set of sense definitions associated with a wordform $w$ in the OED, and $\mathcal{S}_\textrm{PWN}^{w}\subseteq\mathcal{S}$ denote its senses in the PWN. 
Each sense in the OED is part of a lemma, $l\in\mathcal{L}$, which can be recovered trivially; we denote the function for doing so $\lemma_\textrm{OED}^w: \mathcal{S}_\textrm{OED}^{w}\mapsto \mathcal{L}$. 
Our goal is to also map each sense from the PWN to a one of these lemmas, i.e.\ to construct a function, $\lemma_\textrm{PWN}^w: \mathcal{S}_\textrm{{PWN}}^w \mapsto \mathcal{L}$.

No training data for this task exists, so we experiment with simple unsupervised methods. 
Let $\similar$ be a function which takes a pair of definitions, one from each dictionary, and returns a measure of their similarity, $\similar:\mathcal{S}_\textrm{PWN}^{w} \times \mathcal{S}_\textrm{OED}^{w} \mapsto \mathbb{R}$. 
For a particular PWN sense, $s\in\mathcal{S}_\textrm{PWN}^{w}$, these unsupervised models assign the sense to the lemma of the most similar OED sense:\begin{equation} \label{eq:group-assign}
    \lemma_\textrm{PWN}^w(s)= \lemma_\textrm{OED}^w\!\! \bigg ( \!\! \argmax_{~~s'\in \mathcal{S}_\textrm{OED}^{w}}  \similar (s, s') \! \bigg ) \!
\end{equation}

Our methods vary, then, in how they define $\similar$. 
We experiment with very simple approaches, which compute similarity by comparing two definition embeddings. 
Let $\emb$ be a function that produces a \mbox{$d$-dimensional} sentence embedding of a given definition, ${\emb:\mathcal{S}\mapsto \mathbb{N}^d}$. 
Additionally, let $\metric$ be a function which compares two definition embeddings and returns a similarity rating, ${\metric:\mathbb{N}^d\times\mathbb{N}^d\mapsto \mathbb{R}}$. 
We can then express $\similar$ in terms of these functions:
\begin{equation}\label{eq:sim}
    \similar(s, s')= \metric(\emb(s), \emb(s'))
\end{equation}

This formulation allows us to experiment with a variety of different implementations of each of these functions, which we detail in \cref{sec:exp-setup}.

\section{Evaluation}\label{sec:eval}

All of our models are unsupervised, and parameter-free. Each model makes a prediction for each PWN sense in the evaluation data in terms of which lemma in the OED it belongs to. 
In this section, we evaluate how well they do so.

\subsection{Experimental Setup} \label{sec:exp-setup}

\paragraph{Data} To evaluate our models, we use the data we collected in \cref{sec:eval-data}, which consists of $196$ word--part-of-speech combinations, covering $554$ PWN senses. When we evaluate the lemma assignments, we analyse all $554$ senses, for an accurate idea of how the model will perform on the real data (and therefore include senses which were not assigned to a lemma, which the models will necessarily label incorrectly). When we evaluate the sense assignments, however, we filter out all the senses which were not assigned to a sense, leaving $497$ senses. 

\paragraph{Models} Our model formulation centres around a similarity function, \cref{eq:sim}, which has two main components, $\emb$ and $\metric$.
For $\emb$, we experiment with four different sentence embedding models. 
\defn{GloVe} \cite{pennington-etal-2014-glove} is a static embedding technique, which learns to approximate a collocation matrix.
\defn{RoBERTa} \cite{liu2019roberta} is a variant of BERT \cite{devlin-etal-2019-bert}, a Transformer model \cite{vaswani2017attention} which was trained on a masked language modelling objective. 
For both of these embedding spaces, the sentence embedding is taken as the mean of all the token embeddings.
The next two models, \defn{MPNet} \cite{song2020mpnet} and \defn{Sentence-T5} \cite{ni2021sentence-t5}, however, were designed explicitly to produce quality sentence representations. 
MPNet was trained on a variety of tasks for all-round performance, while Sentence-T5 was trained on sentence similarity tasks in particular. 
For all of these sentence embedding models, we use the implementations in the Sentence Transformers Python library \cite{reimers-gurevych-2019-sentence}; where multiple versions are present, we use the largest available. 
The dimensionalities ($d$) of these model's representations are detailed in \cref{tab:embs}.
Each of these embedding spaces might suit different similarity metrics, so for $\metric$, we experiment with dot product, cosine similarity, and Euclidean distance.\footnote{Since Euclidean distance is highest for two senses which are the least similar, we take its negation.} 
Results presented are from whichever similarity metric attained the highest results (in all cases it was dot product).

\paragraph{Baselines} 
We experiment with three baselines. 
As a lower bound for the task, the \defn{random} baseline assigns each sense to a random lemma for a particular word with uniform probability. 
Because some lemmas have more senses than others in the OED, we compute another baseline which assigns each sense to whichever lemma for the word has the \defn{most} OED senses. 
Finally, following \newcite{navigli-2006-meaningful}), we reimplement the LESK algorithm \cite{lesk1986lesk}. The \defn{LESK} baseline calculates the similarity between two definitions, $s$ and $s'$, as the fraction of the shortest definition's lemmas which are in both string definitions:
\begin{equation}
    \similar(s,s') = \frac{|\bow(s)\cap\bow(s')|}{\min(|\bow(s)|,|\bow(s')|)}
\end{equation}
where $\bow$ (bag-of-words) returns the set of lemmas in a given definition. This implementation of $\similar$ is used to find lemma assignments using the same algorithm as the other models, \cref{eq:group-assign}.
To tokenise the definitions and to lemmatise the tokens, we use the word tokeniser and WordNet lemmatiser from NLTK \cite{bird2009natural}. 
We additionally filter out stop words and punctuation, also using the NLTK list for stop words.

\paragraph{Metrics} 
To evaluate the quality of the lemma assignments, we compute accuracy and the F1-score (macro-averaged over the lemmas).
Finding the system that performs best at this level is the core interest in this paper. 
What is important is that a system maps each PWN sense to the correct lemma, which it can do successfully by mapping it to \emph{any} OED sense of that lemma; even if it managed to additionally guess the finer-grained OED sense, this would only be of secondary interest to us. 
However, we are in a situation where we can report performance at a finer granularity because each model internally predicts a fine-grained OED sense. 
We therefore additionally report F1-score and accuracy of these sense assignments (macro-averaged over OED senses).

\paragraph{Significance Testing} 
We use a two-tailed Monte Carlo permutation test at significance level $\alpha=0.01$, with $r=10{,}000$ permutations.

\begin{table}[t]
    \centering
    \begin{tabular}{lr} \toprule
         \textbf{Name} &  $\bm{d}$ \\ \midrule
         GloVe &  $300$ \\ 
         RoBERTa &  $1{,}024$ \\
         MPNet &  $768$ \\
         Sentence-T5  & $768$ \\\bottomrule
    \end{tabular}
    \caption{Sentence embedding dimensionalities}
    \label{tab:embs}
\end{table}

\begin{table*}[t!]
\begin{small}
    \centering
    \begin{tabular}{lcccc} \toprule
         \textbf{Model} & \multicolumn{2}{c}{\textbf{Lemma Assignments}} & \multicolumn{2}{c}{\textbf{Sense Assignments}} \\
          & {Accuracy} &  {F1-score} & {Accuracy} & {F1-Score} \\ \midrule
GloVe & $.94$ & $.93$ & $.71$ & $.70$ \\
MPNet & $.94$ & $.95$ & $.76$ & $.75$ \\
RoBERTa & $.95$ & $.95$ & $.72$ & $.71$ \\
Sentence-T5 & $\bm{.97}$ & $\bm{.97}$ & $\bm{.84}$ & $\bm{.84}$ \\ \midrule
LESK & $.88$ & $.88$ & $.65$ & $.63$ \\
most & $.73$ & $.68$ & N/A & N/A \\
random & $.47$ & $.50$ & N/A & N/A \\ \bottomrule
    \end{tabular}
    \caption{Results}
    \label{tab:results}
    \end{small}
\end{table*}

\subsection{Results}

\cref{tab:results} shows our results. 
Two of the baselines, random and majority, only make lemma assignments, and so we cannot evaluate them at the sense level.

The best performing model overall used the Sentence-T5 embedding space. 
Despite the simplicity of this approach, it attained an F1-score of $0.97$ in the lemma assignment task, the main focus of this work. This was significantly better than all the baselines, and also significantly better than GloVe, the only non-Transformer embedding space. 
Numerically, the difference in the lemma scores was small: GloVe embeddings achieved $.93$ F1, only $.04$ less than Sentence-T5.

In the evaluation data we collected, $1.3\%$ of senses were not assigned to a lemma (see \cref{sec:eval-data}). 
Our model necessarily gets all of these wrong (it has no way of leaving senses unassigned), meaning the highest accuracy it could theoretically attain would be $.98$---only $.01$ higher than it achieves.
For our purposes, that it erroneously assigns these senses is not an issue:
as mentioned above (\cref{sec:coarsening}), because we are interested in research into polysemy and homonymy, we opt to overestimate polysemy and underestimate homonymy, rather than vice versa. This is the effect which this will have.

The best model at predicting the sense-to-sense mapping also used the Sentence-T5 embedding space, but the quality of the mapping was not as high as its sense-to-lemma mapping, attaining an F1 of $.84$. This result is significantly better than not only GloVe, but also both other Transformer models.
The numerical difference between the models is also more pronounced. GloVe attained $.70$ F1, which is $.14$ behind the best Transformer model, and only $.07$ above the LESK approach. 

\section{Final Annotation Layer}

\begin{table*}
    \centering
    \resizebox{\textwidth}{!}{
    \begin{tabular}{lccccccccc}
    \toprule
\textbf{POS} & \multirow{2}{*}{\parbox{1.3cm}{\textbf{\# Words in PWN}}} & \multirow{2}{*}{\parbox{1.3cm}{\textbf{\# Also in OED}}} & \multicolumn{3}{c}{\textbf{\# Homonymous in the OED}} & &  \multicolumn{3}{c}{\textbf{\# Homonymous in the PWN}} \\
& & & \textit{between-POS} & \textit{within-POS} & \textit{raw} & & \textit{between-POS} & \textit{within-POS} & \textit{raw} \\ \midrule
noun & $15{,}019$              & $14{,}228$ &              $806$ & \hphantom{$0{,}$}$849$                & $2{,}830$ & &  $237$ & $244$   &  $794$        \\
verb & \hphantom{$0$}$6{,}226$ & \hphantom{$0$}$5{,}886$ & $237$ & \hphantom{$0{,}$}$310$  &     $1{,}218$    &     & \hphantom{$0$}$50$ & \hphantom{$0$}$56$  & $244$ \\
adj & \hphantom{$0$}$6{,}661$ & \hphantom{$0$}$6{,}115$ & \hphantom{$0$}$75$ & \hphantom{$0{,}0$}$88$ &  \hphantom{$0{,}$}$303$ & & \hphantom{$0$}$17$ & \hphantom{$0$}$17$ & \hphantom{$0$}$54$  \\
adv & \hphantom{$0$}$1{,}037$ & \hphantom{$0{,}0$}$934$ & \hphantom{$00$}$3$ & \hphantom{$0{,}00$}$4$ &  \hphantom{$0{,}0$}$19$ & & \hphantom{$00$}$0$ & \hphantom{$00$}$0$  &  \hphantom{$00$}$2$\\ \midrule
any & $21{,}740$              & $20{,}169$              &  $969$ & $1{,}091$                &    $3{,}420$   &  & $284$ & $297$       & $961$       \\ \bottomrule
    \end{tabular}}
    \caption{Final annotation layer statistics}
    \label{tab:statistics}
\end{table*}

Having performed an evaluation of our approach on  a small testset, we now present details for the entirety of the PWN. 
We use the highest-performing model from our evaluation, which was based on the Sentence-T5 \cite{ni2021sentence-t5} embedding space, and used the dot product to compare embeddings. 

\subsection{Between-POS v.\ Within-POS}

We compute two distinct annotation layer variants, which we term  {between-POS} and {within-POS}. 

The OED is an etymological lexicon, and as such it can identify when two lemmas of the same wordform, but with different parts-of-speech, are derived from each other (this process is called zero-derivation). 
For example, as a verb, to \word{tango} is to perform a particular dance, and as a noun, a \word{tango} is that dance.
In the \defn{between-POS} homonymy annotation layer, we preserve this information, by applying our homonymy identification procedure (\cref{sec:coarsening}) to all the senses of a word at the same time, regardless of their part-of-speech. 

This approach has one drawback. 
As mentioned above, the OED does not have complete information about all senses' etymologies. 
Sometimes, a sense might be labelled with less specific information than another, or might have unknown etymology.
When a wordform had a sense with unknown etymology, we assumed that no homonymy was present, i.e.\ that all the wordform's senses were polysemous. 
This is to reduce the chance of erroneously labelling instances of polysemy as homonymy.
However, in cases where a sense has unknown etymology, there is a chance that we incorrectly treat instances of homonymy as polysemy, an error which we would also like to minimise.

The more senses a wordform has, the more likely it is to have a sense with missing information, which may mean that it is incorrectly treated.
In the \defn{within-POS} layer, when applying our homonymy identification procedure, we treat the senses of each part-of-speech individually. 
This reduces the chance that a sense will be included which lacks etymology information, and so lowers the chance of missing instances of homonymy. 
However, this comes at the price of losing the alignment between different parts-of-speech. 

In both the between-POS and within-POS variants, we exclude OED senses which were not part of the alignment. 
In other words, we first compute the alignment between the PWN and the OED, and then apply our homonymy identification procedure to only the OED senses which are part of the alignment.
This is to minimise the unwanted effects of senses with unknown etymology as much as possible, for both variants.

\subsection{Analysis}

Statistics for the two variants of our annotation layer are presented in \cref{tab:statistics}. 
We additionally report counts using out-of-the-box lemmas from the OED, without  any of the processing in \cref{sec:coarsening}; reported as \defn{raw}. This should give an idea of the number of exclusions resulting from our homonymy identification process.

There are a total of $21{,}740$ words which have multiple senses in the PWN.\footnote{We exclude all wordforms which are not lower case or which include spaces; this removes proper nouns and compound nouns, because these are not included in the OED.} 
Of those, $20{,}169$ ($93\%$) have corresponding entries in the OED. 

Using the within-POS variant, $1{,}091$ wordforms are found to exhibit homonymy.\footnote{Note that for the within-POS variant, the `any' part-of-speech row in \cref{tab:statistics} does not correspond to a simple summation of the statistics for each part-of-speech, because this would count any wordform which is homonymous in two or more different parts-of-speech multiple times.} 
As expected, fewer were found using the between-POS variant ($969$, a reduction of $11\%$).
These numbers represent the maximum number of wordforms in the PWN which our method can identify as exhibiting homonymy. 
Of these, with the within-POS variant we identified $297$ homonymous wordforms in the PWN ($27\%$ of those in the OED), which are associated with a total of $2{,}139$ senses in the PWN (full list of words in \cref{sec:list}). 
With the between-POS variant we identified $284$ wordforms. 
The fact that only a fraction of homonymous wordforms in the OED were also homoynmous in the PWN is unsurprising. 
The OED is an an etymological dictionary, which will contain senses which are no longer used. 
On the other hand, the PWN is a contemporary dictionary, which will not contain archaic instances of homonymy.

Clear-cut cases of homonymy are less numerable than we might expect ($279$ cases; `any' under within-POS in \cref{tab:statistics}). 
These are the cases where wordforms are associated with meanings which have distinct origins and are semantically unrelated. 
But then again, this number represents a lower-bound for the total amount of homonymy in the PWN, as a consequence of our decision to combine lemmas in ambiguous cases. 
An upper-bound (i.e.\ an overestimation of homonymy) is represented by the raw results ($961$ wordforms).
This indicates that between $1.5\%$ and $4.8\%$ of wordforms in the PWN are homonymous (estimated using the wordforms that are in both dictionaries).

\subsection{Release}

We release our code and both variants of our homonymy annotation layer online.\footnote{\url{https://github.com/rowanhm/wordnet-homonymy}} We additionally release a version based on the raw lemma assignments, which will be useful if overestimation of homonymy and underestimation of polysemy is preferred, but we caution that the quality of this data was not investigated in our annotation study.

\section{Conclusion}

We present a new annotation layer for the Princeton WordNet, which splits senses into lemmas, making it possible to distinguish between polysemy and homonymy. 
We use a method which is conservative with respect to homonymy identification (we would rather erroneously label two homonymous senses as polysemous than vice versa, \cref{sec:coarsening}).
Additionally, in contrast to previous work, we use an alignment-based method which will be able to correctly treat figurative polysemy.
We create this annotation layer using a simple method that exploits recent advances in language modelling; although the annotation layer we produce is synthetic, the F1-score that our model attained on a small evaluation set that we produced was $.97$, indicating that it is of high quality.

In future work, we hope to enhance WordNet with more information. 
Lemmas in the OED are annotated with phonetic information; this could be used to infer homophony, which occurs which two unrelated meanings use the same phonetic form (even if they do not necessarily use the same orthographic form). 
An example is the word \word{base}, which is homophonous with the word \word{bass}. 
Additionally, if more complex models could be developed to produce a high quality sense-to-sense mapping to the OED, then we could leverage information the fine-grained senses in the OED contain about the dates of sense emergence, to make WordNet diachronic. 
This would be very useful in the study of language change. 

\section*{Acknowledgements}

We would like to thank the Oxford University Press (OUP) for giving us access to the OED research API, which made this work possible. In particular, we would like to thank Elinor Hawkes from the OUP for helping us with this.

\section{Bibliographical References}\label{reference}

\bibliographystyle{lrec2022-bib}
\bibliography{anthology,bibliography}

\begin{thebibliography}{}

\bibitem[\protect\citename{Ahmadi \bgroup et al.\egroup
  }2020]{ahmadi-etal-2020-multilingual}
Ahmadi, S., McCrae, J.~P., Nimb, S., Khan, F., Monachini, M., Pedersen, B.,
  Declerck, T., Wissik, T., Bellandi, A., Pisani, I., Troelsg{\aa}rd, T.,
  Olsen, S., Krek, S., Lipp, V., V{\'a}radi, T., Simon, L., Gyorffy, A.,
  Tiberius, C., Schoonheim, T., Ben~Moshe, Y., Rudich, M., Abu~Ahmad, R.,
  Lonke, D., Kovalenko, K., Langemets, M., Kallas, J., Dereza, O., Fransen, T.,
  Cillessen, D., Lindemann, D., Alonso, M., Salgado, A., Luis~Sancho, J.,
  Ure{\~n}a-Ruiz, R.-J., Porta~Zamorano, J., Simov, K., Osenova, P., Kancheva,
  Z., Radev, I., Stankovi{\'c}, R., Perdih, A., and Gabrovsek, D.
\newblock (2020).
\newblock A multilingual evaluation dataset for monolingual word sense
  alignment.
\newblock In {\em Proceedings of the 12th Language Resources and Evaluation
  Conference}, pages 3232--3242, Marseille, France, May. European Language
  Resources Association.

\bibitem[\protect\citename{Calabrese \bgroup et al.\egroup
  }2020]{calabrese-etal-2020-fatality}
Calabrese, A., Bevilacqua, M., and Navigli, R.
\newblock (2020).
\newblock Fatality killed the cat or: {B}abel{P}ic, a multimodal dataset for
  non-concrete concepts.
\newblock In {\em Proceedings of the 58th Annual Meeting of the Association for
  Computational Linguistics}, pages 4680--4686, Online, July. Association for
  Computational Linguistics.

\bibitem[\protect\citename{de Melo}2014]{de-melo-2014-etymological}
de~Melo, G.
\newblock (2014).
\newblock Etymological {W}ordnet: Tracing the history of words.
\newblock In {\em Proceedings of the Ninth International Conference on Language
  Resources and Evaluation ({LREC}'14)}, pages 1148--1154, Reykjavik, Iceland,
  May. European Language Resources Association (ELRA).

\bibitem[\protect\citename{Deng \bgroup et al.\egroup }2009]{deng2009imagenet}
Deng, J., Dong, W., Socher, R., Li, L.-J., Li, K., and Fei-Fei, L.
\newblock (2009).
\newblock {ImageNet}: A large-scale hierarchical image database.
\newblock In {\em 2009 IEEE Conference on Computer Vision and Pattern
  Recognition}, pages 248--255.

\bibitem[\protect\citename{McCrae \bgroup et al.\egroup
  }2012]{mccrae2012integrating}
McCrae, J., Montiel-Ponsoda, E., and Cimiano, P., (2012).
\newblock {\em Integrating WordNet and Wiktionary with lemon}, pages 25--34.
\newblock Springer Berlin Heidelberg, Berlin, Heidelberg.

\bibitem[\protect\citename{Meyer and
  Gurevych}2011]{meyer-gurevych-2011-psycholinguists}
Meyer, C.~M. and Gurevych, I.
\newblock (2011).
\newblock What psycholinguists know about chemistry: Aligning {W}iktionary and
  {W}ord{N}et for increased domain coverage.
\newblock In {\em Proceedings of 5th International Joint Conference on Natural
  Language Processing}, pages 883--892, Chiang Mai, Thailand, November. Asian
  Federation of Natural Language Processing.

\bibitem[\protect\citename{Miller and
  Gurevych}2014]{miller-gurevych-2014-wordnet}
Miller, T. and Gurevych, I.
\newblock (2014).
\newblock {W}ord{N}et{---}{W}ikipedia{---}{W}iktionary: Construction of a
  three-way alignment.
\newblock In {\em Proceedings of the Ninth International Conference on Language
  Resources and Evaluation ({LREC}'14)}, pages 2094--2100, Reykjavik, Iceland,
  May. European Language Resources Association (ELRA).

\bibitem[\protect\citename{Miller \bgroup et al.\egroup
  }1994]{miller1994semcor}
Miller, G.~A., Chodorow, M., Landes, S., Leacock, C., and Thomas, R.~G.
\newblock (1994).
\newblock Using a semantic concordance for sense identification.
\newblock In {\em Proceedings of the Workshop on Human Language Technology},
  HLT '94, page 240–243, USA. Association for Computational Linguistics.

\bibitem[\protect\citename{Miller}1995]{miller1995wordnet}
Miller, G.~A.
\newblock (1995).
\newblock {WordNet}: A lexical database for {E}nglish.
\newblock {\em Commun. ACM}, 38(11):39--41, November.

\bibitem[\protect\citename{Navigli \bgroup et al.\egroup }2021]{navigli2021ten}
Navigli, R., Bevilacqua, M., Conia, S., Montagnini, D., and Cecconi, F.
\newblock (2021).
\newblock Ten years of {BabelNet}: A survey.
\newblock In Zhi-Hua Zhou, editor, {\em Proceedings of the Thirtieth
  International Joint Conference on Artificial Intelligence, {IJCAI-21}}, pages
  4559--4567. International Joint Conferences on Artificial Intelligence
  Organization, 8.

\bibitem[\protect\citename{Tan and Bond}2011]{tan-bond-2011-building}
Tan, L. and Bond, F.
\newblock (2011).
\newblock Building and annotating the linguistically diverse {NTU}-{MC}
  ({NTU}-multilingual corpus).
\newblock In {\em Proceedings of the 25th Pacific Asia Conference on Language,
  Information and Computation}, pages 362--371, Singapore, December. Institute
  of Digital Enhancement of Cognitive Processing, Waseda University.

\end{thebibliography}


\begin{thebibliography}{}

\bibitem[\protect\citename{Amaro \bgroup et al.\egroup
  }2006]{amaro2006enriching}
Amaro, R., Chaves, R.~P., Marrafa, P., and Mendes, S.
\newblock (2006).
\newblock Enriching {W}ord{N}ets with new relations and with event and argument
  structures.
\newblock In {\em Proceedings of the 7th International Conference on
  Computational Linguistics and Intelligent Text Processing}, CICLing{'}06,
  page 28–40, Berlin, Heidelberg. Springer-Verlag.

\bibitem[\protect\citename{Bird \bgroup et al.\egroup }2009]{bird2009natural}
Bird, S., Klein, E., and Loper, E.
\newblock (2009).
\newblock {\em Natural language processing with {P}ython: Analyzing text with
  the natural language toolkit}.
\newblock O'Reilly Media, Inc.

\bibitem[\protect\citename{Devlin \bgroup et al.\egroup
  }2019]{devlin-etal-2019-bert}
Devlin, J., Chang, M.-W., Lee, K., and Toutanova, K.
\newblock (2019).
\newblock {BERT}: Pre-training of deep bidirectional transformers for language
  understanding.
\newblock In {\em Proceedings of the 2019 Conference of the North {A}merican
  Chapter of the Association for Computational Linguistics: Human Language
  Technologies, Volume 1 (Long and Short Papers)}, pages 4171--4186,
  Minneapolis, Minnesota, June. Association for Computational Linguistics.

\bibitem[\protect\citename{Freihat \bgroup et al.\egroup
  }2013]{freihat2013regular}
Freihat, A.~A., Giunchiglia, F., and Dutta, B.
\newblock (2013).
\newblock Regular polysemy in {WordNet} and pattern based approach.
\newblock {\em International Journal On Advances in Intelligent Systems}, 6.

\bibitem[\protect\citename{Kernerman \bgroup et al.\egroup
  }2020]{globalex-2020-2020}
Ilan Kernerman, et~al., editors.
\newblock (2020).
\newblock {\em Proceedings of the 2020 Globalex Workshop on Linked
  Lexicography}, Marseille, France, May. European Language Resources
  Association.

\bibitem[\protect\citename{Koch}2016]{koch2016meaning}
Koch, P.
\newblock (2016).
\newblock Meaning change and semantic shifts.
\newblock In Päivi Juvonen et~al., editors, {\em The Lexical Typology of
  Semantic Shifts}, chapter~2, pages 21--66. De Gruyter Mouton.

\bibitem[\protect\citename{Lakoff and Johnson}1980]{lakoff1980metaphors}
Lakoff, G. and Johnson, M.
\newblock (1980).
\newblock {\em Metaphors We Live By}.
\newblock University of Chicago Press.

\bibitem[\protect\citename{Lakoff}1987]{lakoff1987women}
Lakoff, G.
\newblock (1987).
\newblock {\em Women, fire, and dangerous things: What categories reveal about
  the mind.}
\newblock University of Chicago Press.

\bibitem[\protect\citename{Lesk}1986]{lesk1986lesk}
Lesk, M.
\newblock (1986).
\newblock Automatic sense disambiguation using machine readable dictionaries:
  How to tell a pine cone from an ice cream cone.
\newblock In {\em Proceedings of the 5th Annual International Conference on
  Systems Documentation}, SIGDOC '86, page 24–26, New York, NY, USA.
  Association for Computing Machinery.

\bibitem[\protect\citename{Liu \bgroup et al.\egroup }2019]{liu2019roberta}
Liu, Y., Ott, M., Goyal, N., Du, J., Joshi, M., Chen, D., Levy, O., Lewis, M.,
  Zettlemoyer, L., and Stoyanov, V.
\newblock (2019).
\newblock {RoBERTa}: {A} robustly optimized {BERT} pretraining approach.
\newblock {\em CoRR}, abs/1907.11692.

\bibitem[\protect\citename{Mendes and Chaves}2001]{mendes2001enriching}
Mendes, S. and Chaves, R.~P.
\newblock (2001).
\newblock Enriching {WordNet} with qualia information.
\newblock In {\em Proceedings of NAACL 2001 Workshop on WordNet and Other
  Lexical Resources}.

\bibitem[\protect\citename{Navigli \bgroup et al.\egroup
  }2007]{navigli-etal-2007-semeval}
Navigli, R., Litkowski, K.~C., and Hargraves, O.
\newblock (2007).
\newblock {S}em{E}val-2007 task 07: Coarse-grained {E}nglish all-words task.
\newblock In {\em Proceedings of the Fourth International Workshop on Semantic
  Evaluations ({S}em{E}val-2007)}, pages 30--35, Prague, Czech Republic, June.
  Association for Computational Linguistics.

\bibitem[\protect\citename{Navigli}2006]{navigli-2006-meaningful}
Navigli, R.
\newblock (2006).
\newblock Meaningful clustering of senses helps boost word sense disambiguation
  performance.
\newblock In {\em Proceedings of the 21st International Conference on
  Computational Linguistics and 44th Annual Meeting of the Association for
  Computational Linguistics}, pages 105--112, Sydney, Australia, July.
  Association for Computational Linguistics.

\bibitem[\protect\citename{Ni \bgroup et al.\egroup }2021]{ni2021sentence-t5}
Ni, J., {\'{A}}brego, G.~H., Constant, N., Ma, J., Hall, K.~B., Cer, D., and
  Yang, Y.
\newblock (2021).
\newblock {Sentence-T5}: Scalable sentence encoders from pre-trained
  text-to-text models.
\newblock {\em CoRR}, abs/2108.08877.

\bibitem[\protect\citename{Pennington \bgroup et al.\egroup
  }2014]{pennington-etal-2014-glove}
Pennington, J., Socher, R., and Manning, C.
\newblock (2014).
\newblock {G}lo{V}e: Global vectors for word representation.
\newblock In {\em Proceedings of the 2014 Conference on Empirical Methods in
  Natural Language Processing ({EMNLP})}, pages 1532--1543, Doha, Qatar,
  October. Association for Computational Linguistics.

\bibitem[\protect\citename{Piantadosi \bgroup et al.\egroup
  }2012]{piantadosi2012communicative}
Piantadosi, S.~T., Tily, H., and Gibson, E.
\newblock (2012).
\newblock The communicative function of ambiguity in language.
\newblock {\em Cognition}, 122(3):280--291.

\bibitem[\protect\citename{Pu{\c{s}}ca{\c{s}}u and
  Mititelu}2008]{puscasu-mititelu-2008-annotation}
Pu{\c{s}}ca{\c{s}}u, G. and Mititelu, V.~B.
\newblock (2008).
\newblock Annotation of {W}ord{N}et verbs with {T}ime{ML} event classes.
\newblock In {\em Proceedings of the Sixth International Conference on Language
  Resources and Evaluation ({LREC}'08)}, Marrakech, Morocco, May. European
  Language Resources Association (ELRA).

\bibitem[\protect\citename{Pustejovsky}1995]{pustejovsky1995generative}
Pustejovsky, J.
\newblock (1995).
\newblock {\em The generative lexicon}.
\newblock MIT Press.

\bibitem[\protect\citename{Reimers and
  Gurevych}2019]{reimers-gurevych-2019-sentence}
Reimers, N. and Gurevych, I.
\newblock (2019).
\newblock Sentence-{BERT}: Sentence embeddings using {S}iamese {BERT}-networks.
\newblock In {\em Proceedings of the 2019 Conference on Empirical Methods in
  Natural Language Processing and the 9th International Joint Conference on
  Natural Language Processing (EMNLP-IJCNLP)}, pages 3982--3992, Hong Kong,
  China, November. Association for Computational Linguistics.

\bibitem[\protect\citename{Song \bgroup et al.\egroup }2020]{song2020mpnet}
Song, K., Tan, X., Qin, T., Lu, J., and Liu, T.-Y.
\newblock (2020).
\newblock {MPNet}: Masked and permuted pre-training for language understanding.
\newblock In H.~Larochelle, et~al., editors, {\em Advances in Neural
  Information Processing Systems}, volume~33, pages 16857--16867. Curran
  Associates, Inc.

\bibitem[\protect\citename{Utt and Pad{\'o}}2011]{utt-pado-2011-ontology}
Utt, J. and Pad{\'o}, S.
\newblock (2011).
\newblock Ontology-based distinction between polysemy and homonymy.
\newblock In {\em Proceedings of the Ninth International Conference on
  Computational Semantics ({IWCS} 2011)}.

\bibitem[\protect\citename{Vaswani \bgroup et al.\egroup
  }2017]{vaswani2017attention}
Vaswani, A., Shazeer, N., Parmar, N., Uszkoreit, J., Jones, L., Gomez, A.~N.,
  Kaiser, L.~u., and Polosukhin, I.
\newblock (2017).
\newblock Attention is all you need.
\newblock In I.~Guyon, et~al., editors, {\em Advances in Neural Information
  Processing Systems}, volume~30. Curran Associates, Inc.

\bibitem[\protect\citename{Veale}2004]{veale-2004-polysemy}
Veale, T.
\newblock (2004).
\newblock Polysemy and category structure in {W}ord{N}et: An evidential
  approach.
\newblock In {\em Proceedings of the Fourth International Conference on
  Language Resources and Evaluation ({LREC}{'}04)}, Lisbon, Portugal, May.
  European Language Resources Association (ELRA).

\end{thebibliography}

\section{Language Resource References}

\bibliographystylelanguageresource{lrec2022-bib}
\bibliographylanguageresource{anthology,bibliography}

\appendix
\section{List of Homonyms in WordNet} \label{sec:list}

The list below contains the $297$ wordforms which are identified as exhibiting homonymy in the PWN. The $13$ wordforms which appear in the within-POS variant but not the between-POS variant are marked with an asterisk: \newline

\word{adder}, \word{agora}, \word{alum}, \word{angle}, \word{apostrophe}, \word{armed}, \word{ass}, \word{ball}, \word{bank}, \word{bard}, \word{bark}, \word{bar}, \word{bath}, \word{batter}, \word{bat}, \word{beat}, \word{bill}, \word{birr}, \word{boil}, \word{bole}, \word{bongo}, \word{boom}*, \word{boss}*, \word{bowl}, \word{boxer}, \word{boxing}, \word{box}, \word{bracer}, \word{buffer}, \word{buff}, \word{bumble}, \word{bust}, \word{butter}, \word{bye}, \word{calf}, \word{canon}, \word{caper}, \word{carbonado}, \word{castor}, \word{cheese}, \word{chela}, \word{chess}, \word{clove}, \word{coma}, \word{compact}, \word{compound}, \word{content}, \word{con}, \word{copper}, \word{corn}, \word{corona}, \word{cosmos}, \word{courser}, \word{cover}, \word{cramp}*, \word{croup}, \word{cube}, \word{curry}, \word{dam}, \word{deuce}, \word{dick}, \word{diet}, \word{ding}, \word{distemper}, \word{dock}, \word{don}, \word{dory}, \word{down}, \word{drill}*, \word{dub}, \word{excise}, \word{fag}*, \word{fan}, \word{fawn}, \word{feller}, \word{fen}, \word{file}, \word{filicide}, \word{filing}, \word{filler}, \word{flag}*, \word{flat}, \word{flicker}, \word{flop}*, \word{flounce}, \word{forte}, \word{fossa}, \word{full}, \word{fuse}, \word{gall}, \word{game}, \word{gauntlet}, \word{genial}, \word{gill}, \word{gin}, \word{gnarl}, \word{gnome}, \word{gobbler}, \word{gobble}, \word{go}, \word{grad}, \word{grate}, \word{grave}, \word{gray}, \word{gum}, \word{gutter}, \word{gyro}, \word{ha-ha}, \word{hack}, \word{hakim}, \word{hash}, \word{hatched}, \word{hatching}, \word{hatch}, \word{hawker}, \word{hobby}, \word{homer}, \word{hood}, \word{house}, \word{hypo}, \word{impress}, \word{indent}, \word{iridic}, \word{jack}, \word{jar}*, \word{jumper}, \word{junk}, \word{key}, \word{khan}, \word{kip}, \word{kit}, \word{krona}, \word{lame}, \word{launch}, \word{laver}, \word{letter}, \word{lien}, \word{limb}, \word{lime}, \word{ling}, \word{lister}, \word{lithic}, \word{lumber}, \word{lunger}, \word{manakin}, \word{mandarin}, \word{mangle}, \word{mare}, \word{mark}, \word{match}, \word{matted}, \word{matting}, \word{mat}, \word{mean}, \word{meter}, \word{metric}, \word{mew}, \word{mil}, \word{miss}, \word{mogul}, \word{molar}, \word{mole}, \word{monstrance}, \word{mood}, \word{mould}, \word{mow}, \word{mummy}, \word{mush}, \word{must}, \word{nag}, \word{nanny}, \word{nap}, \word{net}, \word{nit}, \word{ore}, \word{paddle}, \word{pall}, \word{para}, \word{pass}, \word{patter}, \word{peewee}, \word{periwinkle}, \word{permit}, \word{phone}, \word{pile}, \word{pink}, \word{pipe}, \word{piping}, \word{pix}, \word{plantain}, \word{plash}, \word{plight}, \word{plonk}, \word{plump}, \word{poacher}, \word{poach}, \word{poise}, \word{poker}, \word{poke}, \word{poll}, \word{pom-pom}, \word{pool}, \word{pop}, \word{port}, \word{pot}, \word{psi}, \word{punch}, \word{punter}, \word{pyrene}, \word{pyrrhic}, \word{python}, \word{quack}, \word{quark}, \word{quid}, \word{quint}, \word{quiver}, \word{race}, \word{racy}, \word{rad}, \word{raft}, \word{raised}, \word{ramp}, \word{real}, \word{reef}, \word{rent}, \word{rest}, \word{retort}, \word{rip}*, \word{roach}, \word{rocket}, \word{rocky}, \word{rock}, \word{rook}, \word{root}, \word{round}, \word{router}, \word{rout}, \word{rue}, \word{rush}, \word{sack}*, \word{sake}, \word{salve}, \word{samba}, \word{sampler}, \word{sardine}, \word{scale}, \word{school}, \word{sconce}, \word{scope}, \word{scourer}, \word{scruple}, \word{scuffle}, \word{seal}, \word{seamy}, \word{secrete}, \word{set}, \word{sewer}, \word{shock}, \word{skipper}, \word{slug}*, \word{snarl}, \word{sod}, \word{sol}, \word{soma}, \word{sort}, \word{sound}, \word{spade}, \word{spanker}, \word{spell}*, \word{spike}, \word{stall}*, \word{stater}, \word{stay}, \word{stereo}, \word{still}, \word{stinger}, \word{stoop}, \word{strain}, \word{tack}, \word{talus}, \word{tanka}, \word{telluric}, \word{temple}, \word{test}, \word{tiller}, \word{timber}, \word{toot}, \word{topi}, \word{tower}, \word{tribune}, \word{tuck}, \word{tuna}, \word{unionized}, \word{verse}, \word{viola}, \word{yen}, \word{zip}

\end{document}